# A Model for Interpreting Social Interactions in Local Image Regions

Guy Ben-Yosef[1,2,3], Alon Yachin[2], Shimon Ullman[2,3]

1. Computer Science and Artificial Intelligence Laboratory, Massachusetts Institute of Technology, Cambridge, MA
2. Department of Computer Science, Weizmann Institute of Science, Rohovot, Israel
3. Center for Brains Minds and Machines, Massachusetts Institute of Technology, Cambridge, MA

gby@csail.mit.edu , alon.yachin@weizmann.ac.il , shimon.ullman@weizmann.ac.il

## Abstract

Understanding social interactions (such as 'hug' or 'fight') is a basic and important capacity of the human visual system, but a challenging and still open problem for modeling. In this work we study visual recognition of social interactions, based on small but recognizable local regions. The approach is based on two novel key components: (i) A given social interaction can be recognized reliably from reduced images (called 'minimal images'). (ii) The recognition of a social interaction depends on identifying components and relations within the minimal image (termed 'interpretation'). We show psychophysics data for minimal images and modeling results for their interpretation. We discuss the integration of minimal configurations in recognizing social interactions in a detailed, high-resolution image.

## Introduction

Understanding social interactions is an important capacity of the human visual system, which starts to develop early in life (Hamlin and Wynn 2011, Mascaro and Csibra 2012, Thomsen et al. 2011). A given social interaction (such as 'hug', 'argue', 'help') can appear in highly variable configurations, in terms of the agents' body pose, their relative positions, the configurations of their hands, their face expressions, and more. Such high variability makes the recognition of social interactions in images a challenging and still open problem.

In this work we present a model for the visual recognition of social interactions, based on small but recognizable local regions. The approach is based on two novel key components. First, we show that a given social interaction (e.g. 'hug') can be recognized reliably from reduced images (called 'minimal images') in which variability is greatly reduced (Ullman et al. 2016, Ben-Yosef et al. 2015;2017). Second, the recognition of a social interaction depends on the internal interpretation of the image, namely, the identification of key components and relations within the image, which are uniquely identified at the level of minimal images. This leads to a model in which a given social interaction is recognized using a number of typical minimal configurations, where each one is recognized by a set of key components and relations identified during learning.

## Related Work

Early research on the visual understanding of social interactions is rooted in the field of social and psychological sciences, studying the different types (e.g., Leary's circumplex, Leary 1957), and physical characteristics (e.g., the distance between two individuals, Hall 1966) of social relations. In terms of computational modeling coming from cognitive and machine vision, only a limited number of studies have addressed the problem of recognizing social interactions from images, most of them using spatiotemporal patterns in video sequences, unlike humans who can also reliably perceive social interactions in still images. Early methods for recognizing interactions were based on characterizing low-level visual features in interaction videos (e.g., Patron-Perez et al. 2012). Recent methods have been based on localizing the agents' body pose (Yang et al. 2012), or face pose (Tanisik, Zalluhoglu, and Ikizler-Cinbis 2016), e.g., by deep CNN features, and modeling the distance between agents (Patron-Perez et al. 2012, Yang et al. 2012).

To test the limitations of recognizing social interactions by existing models, we tested classification and interpretation algorithms for interacting agents, using recent computational methods based on deep feed-forward convolutional networks, and fine tuned to the interaction recognition problem. We collected a dataset of images containing interacting and non-interacting agents (e.g., thousands of images of hugging people, fighting people, superiority interactions etc.), which were used for training and testing. The results show that performance for both classification (e.g., by very deep Feed-forward ConvNets such as He et al. 2015) as well as for interpretation (e.g., by CNN-based keypoints localization for human pose estimation, Chen and Yuille 2014) of interacting agents were significantly lower when compared to human performance (e.g., max classification Average Percision was 59%). Fig 1A. shows some of the classification confusions made by the best methods in our experiments.

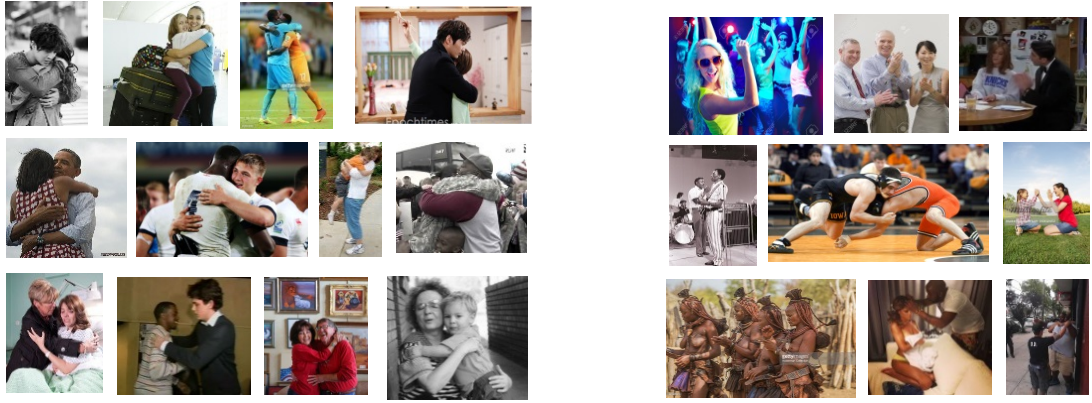

*Figure 1. Samples from our dataset of hugging and non-hugging human agents. The non-hugging images were confused as 'hug' by a fine-tuned DNN-based classifier (He et al. 2015).*

## Minimal Configurations in Social Interaction Images

We describe below the two main components of the model: the identification of basic interaction configurations using minimal images, and the internal interpretation of these configurations.

In the first stage we identified minimal recognizable configurations in social interaction images; these are local image regions in which the interaction type is recognizable, and which further reduction by either size or resolution turns them unrecognizable (Ullman et al. 2016). The search was done using the Amazon Mechanical Turk platform, and usually ended up with several minimal configurations, which were different in the body parts they contained. For example, in a 'hug' image, a minimal configuration could contain the agents' faces and arms, and another only their torsos and arms (without faces, see examples in Fig. 2). A notable characteristic of the minimal interaction images is that small image reduction (i.e., 'sub-minimal' images) can cause large drop in human recognition (Ullman et al. 2016). Examples of minimal configurations found in our experiments are shown in Fig. 2. Examples of minimal and sub-minimal pairs with large drop in recognition are shown in Fig. 3, top and middle rows.

For each minimal configuration, we next identified the components and relations that are required for its recognition. This stage used psychophysical testing of the internal components that humans can recognize in minimal images, varying systematically their size and resolution. The empirical results were analyzed to identify informative components and relations, combining two methods: (i) measuring the drop in recognition between minimal and sub-minimal images caused by a given component or relations and (ii) measuring the contribution of each component or relation, when incorporated in the model, to the algorithm's performance.

For identifying relations and components from minimal and sub-minimal pairs, a minimal image was compared with its similar, but unrecognizable sub-minimal image, to identify either a missing component (e.g., a contour), or a relation between component (e.g., a hand of one agent touches the back of another agent), which were present in the minimal image but not in its sub-minimal configuration. Examples are illustrated in Fig. 3, where pairs of minimal vs. sub-minimal configurations are shown (top and middle rows), along with components or relations that humans can recognize in minimal image, but are not present or satisfied in the sub-minimal configuration (bottom row). The missing component or relation may not be unique; in such cases, we evaluated a number of alternatives.

The bottom row of Fig. 3 includes examples for the most informative features and relations: the existence of arm contours (Fig. 3A-B), the relative location of a palm of one agent to the body of the other agent (Fig. 3C), and the presence of back contours (orange contour in Fig. 3D). Other informative and interesting features and properties were the configuration of a palm (whether it is open or close), and the accurate boundaries of face parts (e.g., the mouth and lips) that carry information regarding relevant face expression.

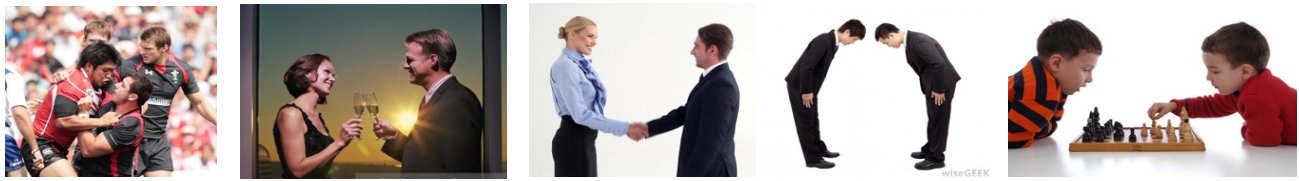
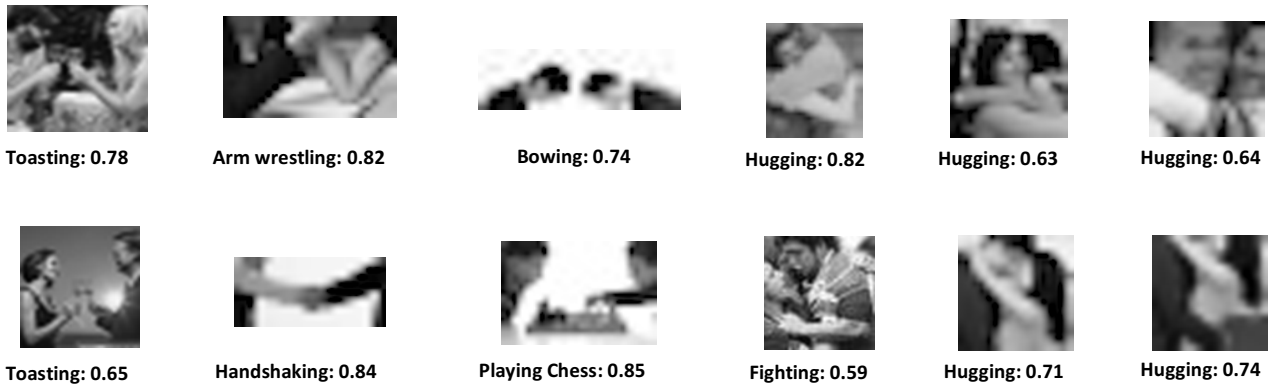

*Figure 2. Social interaction images (**in A**), and minimal configurations found in them and in other images. (**in B**, recognition rate by Mechanical Turk users is on bottom).*

Additional informative features were identified and included in the model.

## Interpretation of Minimal Configurations

The informative components and relations identified in the previous stage, were next incorporated in a computational model for interpreting social interactions, using a structured learning framework (Ben-Yosef et al. 2017). The model proceeds by identifying in an image a configuration of components, which is consistent in terms of the inter-relations with the relations specified by the learned model. The model output for a given new minimal image is an assignment of points, contours, and regions in the image to the various internal components that humans can recognize in the image. For examples, contours are assigned to an agent's arm, and point-features to the eyes.

The interpretation process starts with a candidate region for the social interaction class (e.g., that it contains a 'hug'). The process then uses a learned model of the region's structure to identify within the region the structure that best approximates the learned one. This process has two main stages. The first is to search for local primitives, namely point, contours, and regions in the image to serve as potential candidates for the different components of the expected structure. The second stage searches for a configuration of the components that best matches the learned structure. Examples of the model's interpretation results are in Fig. 4. The results show good agreement with humans' ability to interpret similar images (average of 0.61 Jaccard correspondence, see Tan, Steinbach, and Kumar 2006 for the use of this measure).

For recognizing social interactions in a detailed, high resolution image, we initially detect the minimal configurations contained in the image, for example, configurations containing the face and arm, or the back and palm, etc., and produce interpretations for all these configurations. Recognition of the internal components, and the relations between components, can use at this stage all the available information in the full input image. If the input image contains more than a single minimal configuration, results from the different configurations can next be combined by the model to produce the final interpretation.

## Acknowledgments

This work was supported by ERC Advanced Grant "Digital Baby", the EU's Horizon 2020 research and innovation program under grant agreement No. 720270, Israeli Science Foundation grant 320/16, and the Center for Brains, Minds and Machines, funded by NSF Science and Technology Centers Award CCF-1231216.

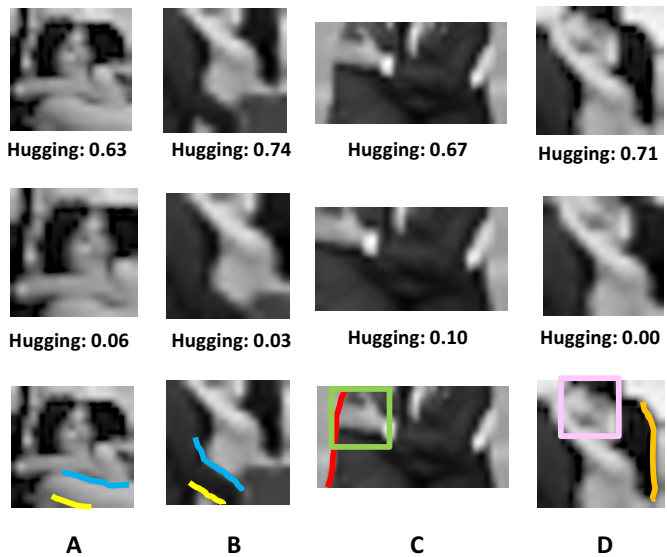

*Figure 3.* Minimal images (top row), sub-minimal images (middle row), and components and relations that can be critical for recognition and interpretation (bottom row). Such components and relations include the arm contours (in A,B), the hand configuration (in C), and back contours (in C,D). The relative weight of the extracted features to the interpretation process was subsequently measured by the model.

# References

Ben-Yosef, G., Assif, L., and Ullman, S. 2017. Full interpretation of minimal images. *Cognition* (in press), volume 171, pp. 65-84. https://doi.org/10.1016/j.cognition.2017.10.006.

Ben-Yosef, G., Assif, L., Harari, D., and Ullman, S. 2015. A model for full local image interpretation. In *Proceedings of the 37th Annual Meeting of the Cognitive Science Society,* 220-225.

Chen, X., and Yuille, A. L. 2014. Articulated pose estimation by a graphical model with image dependent pairwise relations. In *Advances in Neural Information Processing Systems,* 1736-1744.

Hall, E. T. 1966. *The hidden dimension*. New York: Doubleday

Hamlin, J. K., and Wynn, K. 2011. Young infants prefer prosocial to antisocial others. *Cognitive development*, 26(1): 30-39.

He, K., Zhang, X., Ren, S., and Sun, J. 2015. Deep residual learning for image recognition. In *proceedings of the IEEE International Conference on Computer Vision,* 770-778.

Leary, T. 1957. *Interpersonal Diagnosis of Personality*. New York: Ronald Press

Mascaro, O., & Csibra, G. 2012. Representation of stable social dominance relations by human infants. *Proceedings of the National Academy of Sciences*, 109(18), 6862-6867.

Patron-Perez, A., Marszalek, M., Reid, I., and Zisserman, A. 2012. Structured learning of human interactions in TV shows. *IEEE Transactions on Pattern Analysis and Machine Intelligence*, 34(12): 2441-2453.

Tanisik, G., Zalluhoglu, C., and Ikizler-Cinbis, N. 2016. Facial Descriptors for human interaction recognition in still Images. *Pattern Recognition Letters*. 73: 44-51.

Tan, P. N., Steinbach, M., and Kumar, V. 2006. *Introduction to data mining* (Vol. 1). Boston:Pearson Addison Wesley

Thomsen, L., Frankenhuis, W. E., Ingold-Smith, M., and Carey, S. 2011. Big and mighty: Preverbal infants mentally represent social dominance. *science*, 331(6016), 477-480.

Ullman, S., Assif, L., Fetaya, E., and Harari, D. 2016. Atoms of recognition in human and computer vision. *Proceedings of the National Academy of Sciences*, 113(10): 2744-2749.

Yang, Y., Baker, S., Kannan, A., and Ramanan, D. 2012. Recognizing proxemics in personal photos. In *Proceedings of the IEEE Conference on Computer Vision and Pattern Recognition*, 3522-3529.

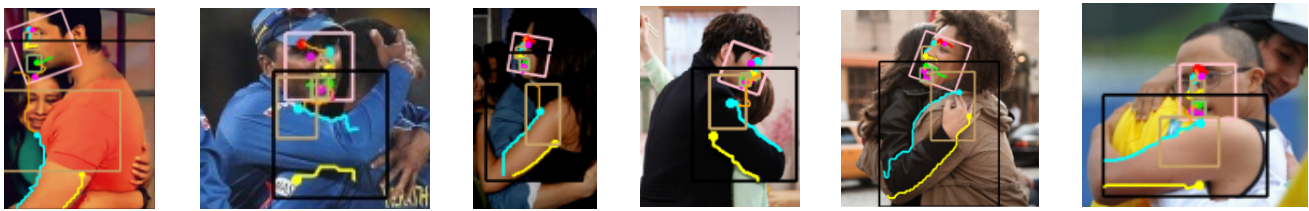

*Figure 4.* Interpretation results of our model for local configurations of hugging images. The model learns point, contour, and region components, and relations between components, from examples of minimal social interaction images. It then can identify these components in novel images.